\documentclass[journal]{IEEEtran}

\ifCLASSINFOpdf
\else
   \usepackage[dvips]{graphicx}
\fi
\usepackage{url}
\usepackage{graphicx}
\usepackage{times}
\usepackage{epsfig}
\usepackage{amsmath}
\usepackage{amssymb}
\usepackage{hhline}

\usepackage{subfig}
\usepackage{booktabs}
\usepackage{float}
\usepackage[ruled, linesnumbered]{algorithm2e} 
\usepackage{amsthm} 

\usepackage{makecell,rotating}
\usepackage{color,xcolor}
\usepackage{multirow}
\usepackage{capt-of,etoolbox}
\usepackage{caption}
\usepackage{mathtools}
\usepackage[colorlinks,linkcolor=blue]{hyperref}

\begin{document}

\title{Human Instance Segmentation and Tracking via Data Association and Single-stage Detector}

\author{Lu~Cheng,~
	     Mingbo~Zhao,~\IEEEmembership{Senior~Member,~IEEE}
		
\thanks{Manuscript received Feb. 19, 2021; revised August 26, 2021. }
\thanks{L. Cheng and M. Zhao are with the School of Information Science and Technology,
Donghua University, Shanghai, China. (e-mail: mzhao4@dhu.edu.cn).}
}

\markboth{Journal of \LaTeX\ Class Files, Vol. 14, No. 8, August 2015}
{Shell \MakeLowercase{\textit{et al.}}: Bare Demo of IEEEtran.cls for IEEE Journals}
\maketitle

\begin{abstract}
Human video instance segmentation plays an important role in computer understanding of human activities and is widely used in video processing, video surveillance, and human modeling in virtual reality. Most current VIS methods are based on Mask-RCNN framework, where the target appearance and motion information for data matching will increase computational cost and have an impact on segmentation real-time performance; on the other hand, the existing datasets for VIS focus less on all the people appearing in the video. In this paper, to solve the problems, we develop a new method for human video instance segmentation based on single-stage detector. To tracking the instance across the video, we have adopted data association strategy for matching the same instance in the video sequence, where we jointly learn target instance appearances and their affinities in a pair of video frames in an end-to-end fashion. We have also adopted the centroid sampling strategy for enhancing the embedding extraction ability of instance, which is to bias the instance position to the inside of each instance mask with heavy overlap condition. As a result, even there exists a sudden change in the character activity, the instance position will not move out of the mask, so that the problem that the same instance is represented by two different instances can be alleviated. Finally, we collect PVIS dataset by assembling several video instance segmentation datasets to fill the gap of the current lack of datasets dedicated to human video segmentation. Extensive simulations based on such dataset has been conduct. Simulation results verify the effectiveness and efficiency of the proposed work.
\end{abstract}


\IEEEpeerreviewmaketitle

\section{Introduction}

\IEEEPARstart{I}{n} the field of computer vision research area, instance segmentation is a typical challenge to perform pixel-level segmentation of different instances. Among different instance segmentation learning tasks, human instance segmentation plays an important role in computer understanding of human activities and is widely used in video processing, video surveillance, and human modeling in virtual reality. On the other hand, with the rapid growth of Internet and mobile terminals there is an exponential increase in video data, such as short videos, pedestrian videos in in mobile entertainment, surveillance or autonomous driving system, most of which are human-centred. To effectively analyze human activities in a certain video, an elementary task is to segment and track the human instances in the entire video sequences so as to accurately grasp the main dynamic information of human instances. As a result, human video instance segmentation (HVIS), as a more challenging learning task instead of only segmenting human instance in a single image, has gained significantly increased interests due to its practical utilization.  

In general, given a video sequence, the goal of human video instance segmentation \cite{yu2021real} is to realize pixel-level segmentation for each human instance in every single frame, as well as to track the human instance across the video frames \cite{smeulders2013visual}. From this view of point, human video instance segmentation is also an extended learning task for object tracking (OT). Similar to the OT, the dataset for video instance segmentation \cite{caelles20192019,voigtlaender2019mots,yang2019video}, have some key problems, such as motion blur, occlusion or motion deformation, that will affect the image quality. This will fail to meet the requirement of VIS, i.e. high segmentation accuracy, good association effect and high real-time performance. Many work have been proposed to handle the above problems and enhance the VIS performance during the past decades \cite{caelles20192019,voigtlaender2019mots,yang2019video}. In practice, based on whether the masks of instances are given in the first frame, the video instance segmentation methods can be generally divided into two categories: mask propagation-based methods and the tracking-by-detection ones.
 
The method based on mask propagation \cite{bertasius2020classifying,oh2018fast,oh2019video,voigtlaender2019feelvos} is applicable to the case where the target appears in the first frame of the video. In detail, in mask propagation based methods, the instance mask of the first image frame is firstly obtained by human labelling or a certain instance segmentation method. It then utilizes such information in the first frame as guidance to propagate the instance mask to the subsequent video frames. The mask propagation based methods can handle long-term VIS problem even if the instance in the following frames is sheltered. However, these methods relies on the segmentation accuracy of the first frame. Given the case that a higher quality instance mask cannot be acquired, the subsequent mask propagation will aggravate the mask segmentation error and the quality of video instance segmentation will be seriously affected. Therefore, the masks in the first frame preserved in the memory can better be updated in a certain interval to be consistent with the followed video frame. But this will definitely increase computation cost. 

To the contrast, the methods based on tracking-by-detection paradigm \cite{cao2020sipmask,wang2021end,peng2020chained} works in a short-term way by simultaneously segmenting and tracking the instance in a video sequence. In detail, the tracking-by-detection based methods firstly calculate the instance masks in each video frame via instance segmentation branch of detector and then match the same instance in the video frame by certain data association methods, which include simple IOU matching \cite{bochinski2017high}, feature expression based matching learning, joint learning of detection and matching features \cite{zhang2017survey} et al. The continuous motion trajectory of the instance can thereby be formulated. However, most existing VIS methods are based on Mask-RCNN framework, a typically two-stage detector approach, that firstly generate multiple instance proposals and then conduct feature pooling operations on each proposal for calculating its bounding box, class label, mask as well as embedding for matching the same instance with existing trajectories in a video. A key drawback of these methods is that it requires designing and fine-tuning a separate model for each sub-task. In addition, the end-to-end training \cite{xu2020train} that include target appearance and motion information for data matching will nevertheless increase computational effort and have an impact on segmentation real-time performance. 

In this paper, to handle the above problems, we develop a new method for human video instance segmentation, which is based on SOLO \cite{wang2020solo} framework by considering the case where the target does not appear in the first frame and the requirement of segmentation efficiency. Different from Mask-RCNN based methods \cite{yang2019video}, SOLO method is a one-stage instance segmentation method that does not require proposal generations for dense predictions of boxes and instance masks. In addition, the global mask prediction branch induced by SOLO has the same resolution for the objects of different sizes, hence it exhibits natural advantages to have clearer objects and boundaries. Motivated by this end, we will consider to extending SOLO framework for handling HVIS problem. In detail, to tracking the instance across the video, we have adopted data association strategy for matching the same instance in the video sequence, where we jointly learn target instance appearances and their affinities in a pair of video frames in an end-to-end fashion. Here, we firstly model the instance appearance with a two-stream convolutional network with shared parameters, we then have adopted the centroid sampling strategy for enhancing the embedding extraction ability of instance, which is to bias the instance position to the inside of each instance mask with heavy overlap condition. As a result, even there exists a sudden change in the character activity, the instance position will not move out of the mask, so that the problem that the same instance is represented by two different instances can be alleviated. All these strategies can promote robustness against instance occlusions in the induced model; finally, we collect PVIS dataset by assembling several video instance segmentation datasets to fill the gap of the current lack of datasets dedicated to human video segmentation. Extensive simulations based on such dataset has been conduct. Simulation results verify the effectiveness and efficiency of the proposed work.

The main contributions of the proposed work are as follows:

1) we have developed a new VIS method by incorporating feature extraction and data association components into a single-stage segmentation network SOLO \cite{wang2020solo}, where the target recognition, segmentation, and tracking can be realized simultaneously. The one-stage framework with data association can guarantee the efficiency and accuracy for tracking, while the global mask induced by SOLO can have clearer objects and boundaries since it has the same resolution for the objects of different sizes;  

2) we have adopted a two-stream convolutional network strategy to extract the embeddings and of instance calculate the similarity of each instance between any two frames from the extracted appearance embeddings. We also adopt a centroid sampling strategy for enhancing the embedding extraction ability to deal with heavy overlap condition and guarantee the instance embeddings will not be treated as two given there is sudden change in the character activity. The occlusion problem in the tracking process is thereby improved, as is the presence of ID switches of the same instance at different scales.

3) The human video segmentation dataset PVIS is constructed by assembling several video instance segmentation datasets to fill the gap of the current lack of datasets dedicated to human video segmentation for evaluation of algorithm accuracy and efficiency.

The following work is organized as follows: in Section II, we briefly review some related work about VIS, tracking-by-detection methods and mask propagation methods; in Section III, we will provide detailed description for illustrating the proposed work for human VIS. Extensive simulations are conducted in Section IV and final conclusions are drawn in Section V.
\section{Related Work}
\subsection{Video Instance Segmentation}
Instance segmentation is an extension to the semantic segmentation, which is a challenging task that requires both pixel-level and instance-level mask prediction. Many instance segmentation methods have been proposed during the past decades. The traditional two-stage approach Mask-RCNN \cite{he2017mask} is to directly extend the Faster-RCNN \cite{ren2015faster} by adding a segmentation branch for segmenting instances in a priori bounding box. Single-stage approaches that reduce feature alignments have emerged in recent years. For example, YOLACT \cite{bolya2019yolact} builds prototype masks of image size based on FCN, predicts a set of coefficients, and then utilizes a linear combination of coefficients to generate instance masks. SOLO \cite{wang2020solo} transforms the instance segmentation problem into a single-stage classification problem by assigning a categories to each pixel based on the location and size of each instance. This technique, which adopts a simple framework with good performance, does not require an anchor box or a bounding box.

Video instance segmentation, in contrast to instance segmentation, requires not only the instance segmentation of a single frame, but also the consideration of instance relationship between different frames. MaskTrack-RCNN \cite{yang2019video} first proposed the task of video instance segmentation and provided a large video dataset called YouTube-VIS. This method extends Mask-RCNN \cite{he2017mask} and adds an identity embedding branch to handle the instance association task. However, the approach sole considers appearance features and does not take advantage of video sequences temporal information. To handle this problem, CompFeat \cite{fu2020compfeat} refines frame-level and object-level features using temporal and spatial contextual information. VisTR \cite{wang2021end} implements a query-based end-to-end approach using DETR \cite{carion2020end}.

\subsection{Tracking-by-Detection Paradigm}
The VIS can be divided into two categories: the tracking-by-detection based methods and mask propagation-based methods. In tracking-by-detection based methods, the instance segmentation approach is firstly used to predict the instance mask for each frame. Then, the appearance features of each instance are retrieved via a re-identification model, and the motion trajectory is generated by computing the appearance and motion affinity. Here, a key technology for tracking by detection methods is the data association for calculating the affinity of the same instance between different video frames. SORT \cite{bewley2016simple} based methods are typically utilized for data association by utilizing the Kalman filter \cite{maybeck1990kalman} and the Hungarian matching algorithm \cite{kuhn1955hungarian}. However, these methods that execute the identification and tracking processes independently are computationally complex and perform poorly in real-time. Due to the advancement of multi-task learning \cite{kendall2018multi}, joint detection and tracking approaches \cite{chaabane2021deft,wang2020towards,zhang2021fairmot} are becoming more popular due to their efficiency. By adding parallel branches to the object detector, these methods are able to forecast an object's tracking offset or re-identify features for data association. \cite{peng2020chained} further fully aggregates object detection, feature extraction, and data association into one model for end-to-end tracking detection.

\subsection{Mask Propagation Methods}
In such approaches, the instance mask of the first frame is obtained as a guide to achieve mask propagation by a semi-supervised video object segmentation method. For example, MaskProp \cite{bertasius2020classifying} includes a mask propagation branch based on Mask-RCNN \cite{he2017mask}, as well as temporal sampling network and high-resolution mask refinement post-processing, to propagate the object instance mask of a video frame to the other frames using deformable convolution \cite{dai2017deformable}. Spatio-temporal memory network (STM) \cite{oh2019video} introduces a memory mechanism \cite{bertasius2018object} to store the information of previous frames and then reads it by querying the frames to extend the matching range to the whole video range. MaskRNN \cite{hu2018maskrnn} captures temporal correlation using a recursive recurrent network and exploits the video's long-term temporal structure for matching. The mask propagation approach can make greater use of the temporal information between video frames and handle smooth motion conditions. However, the propagation effect is hampered when there are fast changes in position or appearance. In addition, these methods also heavily relies on the segmentation accuracy of the first frame. 

\section{The Proposed Method}
\subsection{Overview}
\label{sub: Overview}
\begin{figure*}
\begin{center}
\includegraphics[width=0.9\textwidth]{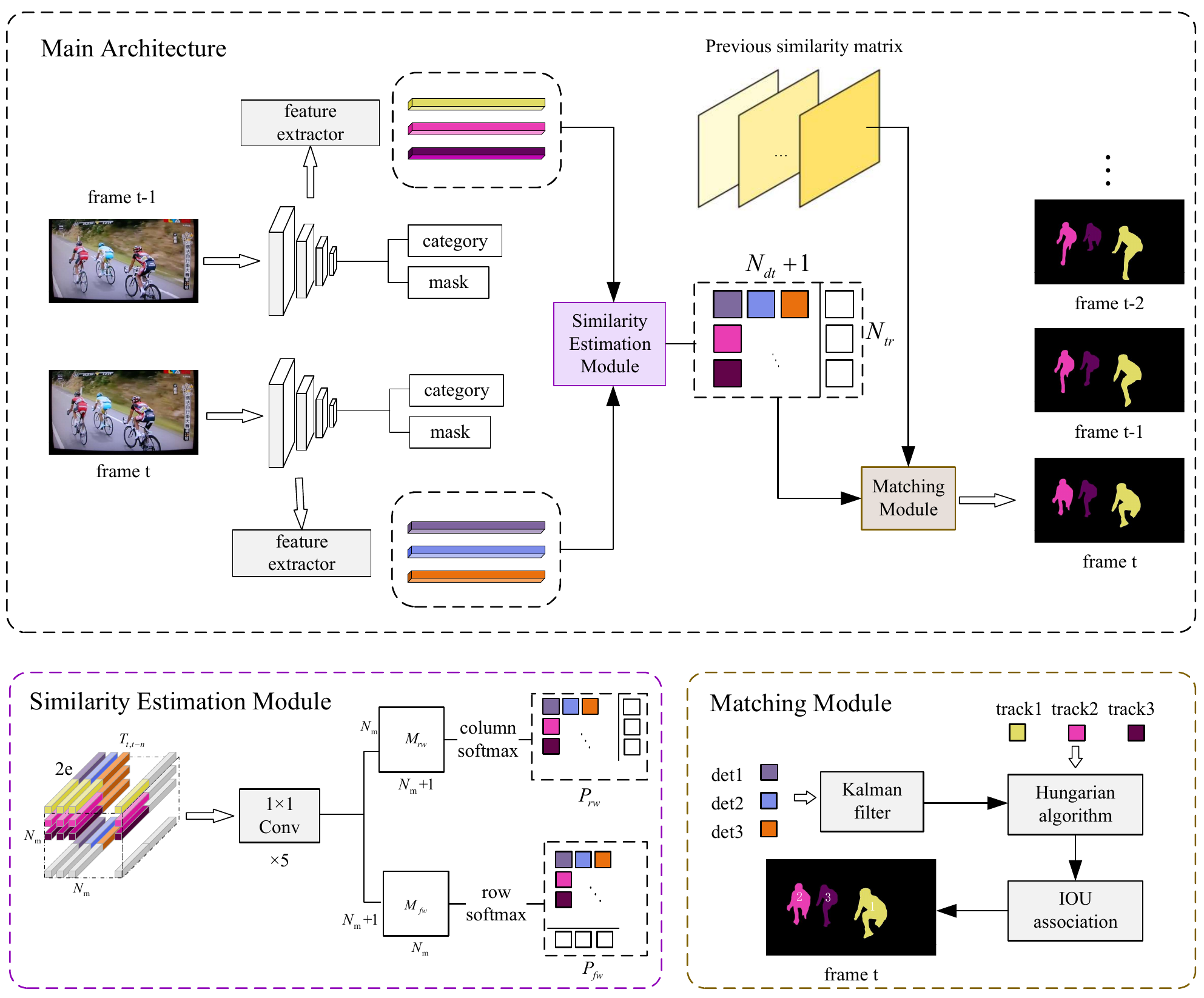}
\end{center}
\caption{The overview of the proposed method. The images are first input to the network. The feature extraction module extracts feature embeddings from the feature layers at various scales of the backbone network to represent the instances. Then the similarity estimation module is used to calculate the probability that different features in two frames belong to the same instance. Finally, in the matching module, the existing instance identity information is matched for the current frame instance based on the stored similarity matrix, which in turn forms the continuous motion trajectory of the instance in the video.}
\label{f1}
\end{figure*}

The article's general aim is to be able to accurately segment all people in a video with complicated backgrounds and shifting scenery, as well as fix the identity assignment of the same person. The specific approach is based on the bottom-up global instance segmentation method SOLO, the overall framework of which is shown in Fig. \ref{f1}. SOLO \cite{wang2020solov2} transforms the instance segmentation problem into a pixel-level classification task. The input image is divided into $S \times S$ grids, and if the center of an instance falls in one of the grids, that grid is responsible for the instance classification and mask prediction tasks. A feature extraction module is added to the backbone network to extract the instances' appearance embedding, and a similarity estimation module is added to correlate the instance embedding between various frames, which performs instance segmentation in whole video.
\subsection{Feature Extraction Module}
The purpose of the feature extraction module is to get feature embeddings to represent instances and to use appearance embeddings to perform data association between video frames. A pair of video frames with n timestamps interval are fed to Siamese network structure ($n\in \left[ 1,\left. {{T}_{m}} \right] \right.$). In the process of video instance segmentation, large scale changes will occur due to the motion of the camera or the person, while target association requires the feature of the instance to have scale invariance. And the segmentation principle of SOLO \cite{wang2020solo} is generating predictions by grid cells according to the masks’ size, and the same instances of different scales correspond to different instances. The embeddings concatenated directly from the FPN structure \cite{lin2017feature} can lead to serious id-switch problems. Instead, a total of ${{N}_{f}}$ feature maps are selected directly on the ResNet-101 \cite{he2016deep} backbone and the instance features at the corresponding locations are extracted. To boost the expressiveness of the instance embedding, feature maps with low resolution but high semantic information and feature maps with poor semantic information but rich spatial information are fused in this process. The Siamese network structure \cite{leal2016learning} also learns the relevant spatio-temporal features, which enhances the robustness of embeddings. In response to computational limitation, $1\times1$ convolutional kernels are used to reduce the dimensionality before extracting the features from the center of instances.

Directly extracting the embedding from the center of the instance bounding box, on the other hand, provides poor discriminating for people of comparable size and substantial overlap, resulting in the same positive sample representing two different people. For this type of fuzzy positive sample problem, the centroid sampling strategy for maximum area contour is utilized, which can bias the sample position to the inside of each instance mask with heavy overlap condition. After a sudden change in the character activity, the sample positions do not move out of the mask, which alleviates the fuzzy positive sample problem. Thus we introduce the instance feature sampling centers:
\begin{equation}
{m_{pq}} = \sum\limits_{y = 1}^{{H_m}} {\sum\limits_{x = 1}^{{W_m}} {{x^p} \cdot {y^q} \cdot c{t_{\max }}(x,y)} }
\label{e1}
\end{equation}
\begin{equation}
{x_c} = \frac{{{m_{10}}}}{{{m_{00}}}},{y_c} = \frac{{{m_{01}}}}{{{m_{00}}}}
\label{e2}
\end{equation}
where ${{W}_{m}}$ and ${{H}_{m}}$ are the feature map dimensions at different scales, $c{t_{\max }}$ is the binary image corresponding to the maximum mask area of the same instance, ${m_{pq}}$ is the $(p+q)$-order moment, and ${({x}_{c},{y}_{c})}$ is the centroid position of the feature to be sampled. The ${{i}_{th}}$ instance's appearance embedding ${{E}_{{{t}_{\text{i}}}}}\text{= }\!\!\{\!\!\text{ }{{E}_{({{x}_{{{1}_{i}}}},{{y}_{{{1}_{i}}}})}}\text{,}{{E}_{({{x}_{{{2}_{i}}}},{{y}_{{{2}_{i}}}})}}\text{,}\cdots \text{,}{{E}_{({{x}_{{{N}_{f}}_{i}}},{{y}_{{{N}_{f}}_{i}}})}}\text{ }\!\!\}\!\!\text{ }\in {{\mathbb{R}}^{e\times 1}}$ at time $t$ was derived by sampling the location of centroid for maximum area contour. To construct the end-to-end training process, the maximum number of instances that can be detected by a single video is fixed to be ${{N}_{m}}$ .Then we obtain the appearance feature matrix ${{E}_{t}}\in {{\mathbb{R}}^{e\times {{N}_{m}}}}$ of the instance in the ${{t}_{th}}$ frame of the video, and similarly ${{E}_{t-n}}\in {{\mathbb{R}}^{e\times {{N}_{m}}}}$ in the $(t-n){{}_{th}}$ frame. 
\subsection{Similarity Estimation Module}
The similarity estimation module calculates the similarity of each instance between the two frames from the extracted appearance embeddings, in other words, it calculates the probability that the appearance embeddings belong to the same instance. The similarity estimation module is based on the structure of DAN \cite{sun2019deep}. Since the number of people that can be detected in the video is fixed at ${{N}_{m}}$, in addition to the actual extracted embeddings, the feature matrix ${{E}_{t}}$ and ${{E}_{t-n}}$ of all instances of the two frames can be obtained by using the all-zero vector to complement to the set maximum number of instances. The feature matrices representing the features of different instances of video frames are combined in pairs, and the corresponding feature embeddings are concatenated to get the tensor ${{T}_{t,t-n}}\in {{\mathbb{R}}^{2e\times {{N}_{m}}\times {{N}_{m}}}}$ to represent all instance matching. In order to intuitively express the appearance similarity of the two instances, the tensor is compressed along the direction of the embedding vector through five $1\times1$ convolution kernels to obtain ${{T}_{t,t-n}}^{\prime }\in {{\mathbb{R}}^{{{N}_{m}}\times {{N}_{m}}}}$. That is to say, the feature vectors of the instance are directly mapped to the similarity matrix of instances. 

Since the tensor ${{T}_{t,t-n}}^{\prime }$ only represents the appearance similarity of the actual instance in frame t and frame $t-n$, it cannot process the instance leaving the video or entering the video between two frames. In this case, adding a vector of size ${{N}_{m}}$ represents the possibility that at most ${{N}_{m}}$ instances will leave or enter the video. Since the entry of a new instance when the ${{t}_{th}}$ frame is associated to an already existing instance in the $(t-n)_{th}$ frame is the same as the departure of an old instance when the $(t-n)_{th}$ frame is associated to an already existing instance in the ${{t}_{th}}$ frame. Therefore, appending a row and column vector respectively represents the probability that the instance leaves the video when frame $t$ is associated with the instance of frame $t-n$ and the probability that the instance enters the video when frame $t$ is associated with the instance of frame $t-n$, thus obtaining ${{M}_{fw}}\in {{\mathbb{R}}^{({{N}_{m}}\text{+}1)\times {{N}_{m}}}}$ and ${{M}_{rv}}\in {{\mathbb{R}}^{{{N}_{m}}\times ({{N}_{m}}+1)}}$.The vector exists in terms of $v\in {{\mathbb{R}}^{{{N}_{m}}}}=\gamma \times 1$, where $\gamma $ is a learnable hyperparameter.

${{M}_{fw}}$ represents the association between the instances of the $(t-n){{}_{th}}$ frame and the ${{t}_{th}}$ frame, which is defined as the forward association. Accordingly, ${{M}_{rv}}$ is defined as reverse association. Then, the ${{m}_{th}}$ line of ${{M}_{fw}}$ represents the association between the instance numbered m in the $(t-n){{}_{th}}$ frame and all instances in the ${{t}_{th}}$ frame. The corresponding matching probability distribution is obtained in each column through the operation of column softmax, and the similarity probability matrix ${{P}_{fw}}\in {{\mathbb{R}}^{({{N}_{m}}+1)\times {{N}_{m}}}}$ represents the association relationship in the form of probability. Similarly, the corresponding matching probability distribution is obtained in each row through the operation of row softmax, and the similarity probability matrix ${{P}_{rv}}\in {{\mathbb{R}}^{{{N}_{m}}\times ({{N}_{m}}+1)}}$ is obtained to represent the association relationship between the instances of frame t and frame $t-n$ in the form of probability.  

According to the above analysis, association process can be divided into forward association and reverse association, and association loss also includes forward loss ${{L}_{fw}}$ and reverse loss ${{L}_{rv}}$. The forward loss ensures the correct association of instances from frame $t-n$ to frame $t$, and the reverse loss guarantees the correct association of instances from frame t to frame $t-n$. At the same time, in order to suppress the association between instances with non-maximum similarity, the non-maximum loss ${{L}_{nm}}$ is added to maximize the actual instance association probability matrix. The final loss is the average of the three parts.  
\begin{equation}
{L_*} = \frac{{\sum\limits_{i = 1}^{row} {\sum\limits_{j = 1}^{col} {\left( {{G_{\rm{*}}}_{(i,j)}( - \log {P_{\rm{*}}}_{(i,j)})} \right)} } }}{{\sum\limits_{i = 1}^{row} {\sum\limits_{j = 1}^{col} {{G_*}_{(i,j)}} } }}
\label{e3}
\end{equation}
\begin{equation}
{L_{nm}} = \frac{{\sum\limits_{i = 1}^{{N_m}} {\sum\limits_{j = 1}^{{N_m}} {\left( {{G_{(i,j)}}( - \log \max ({{P'}_{fw(i,j)}}){{P'}_{rv(i,j)}})} \right)} } }}{{\sum\limits_{i = 1}^{{N_m}} {\sum\limits_{j = 1}^{{N_m}} {{G_{(i,j)}}} } }}
\label{e4}
\end{equation}
\begin{equation}
{L_{match}} = \frac{1}{3}({L_{fw}} + {L_{rv}} + {L_{nm}})
\label{e5}
\end{equation}
where $\text{*}$ represents the forward and reverse directions, $G$ is the actual instance ground truth association matrix, and ${{G}_{\text{*}}}$ is the ground truth association matrix with the added dynamic case. ${{P}_{\text{*}}}^{\prime }$ removes the added dynamic case compared with ${{P}_{\text{*}}}$.
\subsection{Tracking Trajectory Generation}
In order to achieve instance association between frames, after obtaining the feature embeddings of the instance in a single image, features are stored for nodes according to the sequence of corresponding video frames. The similarity between the current frame and the previously stored node features is calculated and stored in corresponding timestamp. The identity of the stored instance is assigned to the detected unknown instance based on the node information, which in turn assigns the same identity to the same instance in the video, forming a continuous tracking trajectory. 

When defining the ${{t}_{th}}$ frame, there are ${{N}_{tr}}$ tracks, corresponding to ${{N}_{tr}}$ instances of assigned identity, and ${{N}_{dt}}$ instances detected. The bidirectional similarity probability matrix ${{P}_{\text{*}}}$ between the current ${{t}_{th}}$ node and each stored node is calculated according to the previous $t-1$ node records. Then the similarity matrix ${{S}_{(t,t-1)}}$ of the instance of frame $t$ and frame $t-1$ is set as follows:
\begin{equation}
{{S}_{(t,t-1)}}=\left\{ \begin{matrix}
   \max ({{{{P}'}}_{fw(x,y)}},{{{{P}'}}_{rv(x,y)}}) & x,y\in [0,{{N}_{m}}]  \\
   {{P}_{fw(i,{{N}_{m}}+1)}} & i\in [0,{{N}_{m}}]  \\
\end{matrix} \right.
\label{e6}
\end{equation}
According to the actual number of detected and tracked instances, ${{{S}'}_{(t,t-1)}}\in {{\mathbb{R}}^{{{N}_{tr}}\times ({{N}_{dt}}+1)}}$ is obtained, where rows represent stored instance identities, columns represent unassigned instances detected by the current frame, and additional columns represent dynamic situations. After obtaining the similarity matrix between the current frame and all previous nodes, the similarity matrix of the latest four nodes was merged, and the median of the four nodes was taken to construct the instance allocation problem of the ${{t}_{th}}$ frame detection result and tracking result. Choosing four nodes can not only collect multi-frame information, reduce the influence of missing or temporary tracking failure on identity association, but also avoid the huge memory problem caused by directly selecting all nodes. At the same time, the selection of median method can eliminate some extreme cases compared with the average method.

The matching process is divided into three steps: features association, IOU association and Kalman filtering. Firstly, the correlation results of motion trend inconsistent with Kalman \cite{maybeck1990kalman} prediction were filtered. The association factor of the feature embeddings association part is the appearance information of the instance. Finally, the Hungarian algorithm \cite{kuhn1955hungarian} is used to maximize the similarity between the current frame and the instance object which has been assigned identity, and then the instance identity and the detected instance are matched one-to-one. The trace and detection instances that fail the first match are associated with IOU. This part applies to the case where the instance moves slowly, and determines whether it belongs to the same instance by calculating the coincidence degree of boundary boxes. Because the usage of a boundary box is not appropriate in some instances involving significant action amplitudes, the mask IOU association is used as follows:
\begin{equation}
	MaskIOU=\frac{Are{{a}_{dt}}\cap Are{{a}_{tr}}}{Are{{a}_{dt}}\cup Are{{a}_{tr}}}
\label{e7}
\end{equation}
where $Are{{a}_{dt}}$ is the number of internal points of the detected instance mask, and the mask indicator is used to further accurately represent the degree of location overlap of the targets. After three steps, one-to-one matching is finally achieved. The detection object that still fails to match is entered into the next match as a new tracking object. Meanwhile, due to the memory limitation, it cannot store all the instance motion trace, and the time $\tau$ limits the maximum storage time. If this limit is exceeded, the trace in the trace sequence is deleted.

\subsection{Training}
During the training, two frames of video images were selected as input pairs, and features were extracted using Siamese network structure. According to the video instance segmentation task, the loss function was divided into detection, segmentation and matching parts, and the specific calculation was as follows: 
\begin{equation}
	L\text{=}\frac{1}{2}(\frac{1}{{{e}^{{{s}_{i}}}}}({{L}^{t-n}}_{det-seg}+{{L}^{t}}_{det-seg})+{{s}_{i}}+\frac{1}{{{e}^{{{s}_{j}}}}}{{L}_{match}}+{{s}_{j}})
\label{e8}
\end{equation}
where ${{s}_{i}}$ and ${{s}_{j}}$ are learnable parameters that can be used to deal with the unpredictability of task loss and lessen the influence on a single task during joint training. In Eq. \eqref{e8}, ${{L}_{det-seg}}$ is obtained by the sum of ${{L}_{cat}}$ and ${{L}_{seg}}$, which respectively represent the classification loss and segmentation loss of \cite{wang2020solo}. And the segmentation loss of two frames of video images needs to be calculated and summed simultaneously.${{L}_{match}}$ is calculated by Eq. \eqref{e3}, Eq. \eqref{e4}, Eq. \eqref{e5}. Since the evaluation scales of segmentation task and matching task are different and final effects depend heavily on the weight of task’s loss, the training effects of both tasks will be influenced by direct summing with identical loss weights. Therefore, multi-task learning \cite{kendall2018multi} is used to automatically balance the training process of loss. 

\section{Experiments}
\subsection{Dataset and Evaluation Methods}
\label{sub: datasets}
\textbf{Dataset:}

To fill the gap of the current lack of datasets dedicated to human video segmentation, we construct a new benchmark dataset  on the basis of the existing video instance segmentation datasets \cite{caelles20192019,yang2019video,fan2019shifting} to train and evaluate the proposed method. The constructed dataset should satisfy three requirements: (1) more accurate annotations; (2) more complex video scenes with a variety of challenging scenes, such as occlusion, motion blur and appearance deformation; (3) avoiding ambiguous annotation in the case of only a few significant people in the crowd.

According to the above dataset requirements, we re-label and screen the existing datasets, including YouTube-VIS \cite{yang2019video}, DAVIS \cite{caelles20192019} and DAVSOD \cite{fan2019shifting}, to extract the human parts and then obtain new datasets for training and evaluation. DAVSOD \cite{fan2019shifting} only marks the prominent objects in the video, while DAVIS \cite{caelles20192019} focuses on the prominent people in the first frame, but not the people appearing in the middle or the situation of crowded people. By comparing the above two datasets, YouTube-VIS \cite{yang2019video} can basically annotate all the people in different periods of videos. Therefore, the new dataset constructed based on YouTube -VIS deletes some video sequences that do not meet the above requirements, and adds some sequences that meet the requirements in other two datasets, where the instance serial number is reallocated. The YouTube-VIS \cite{yang2019video} dataset contains 2.9K videos, 40 common categories, 4883 instances and the corresponding 131K annotated masks. 

As a result, the new proposed VSI dataset (PVIS) contains 1117 videos with 2025 objects in the videos and about 60k annotated instance masks, of which 128 video sequences are used for method evaluation. Since the new dataset is composed of multiple datasets and the interval between adjacent frames is 1-5 FPS, the lower frame rate will lead to the larger inter-frame motion in some fast-moving videos, which also puts forward higher requirements for instance matching.

\textbf{Evaluation Metric}: 

Video instance segmentation is a complex task that integrates detection, segmentation and tracking. HOTA \cite{luiten2021hota} measurement, which can simultaneously measure the effect of detection and instance association, is adopted as the evaluation method. HOTA balances multi-task metrics by calculating geometric average values of detection and associated parts. In case of missed detection instances or error detection, a lower detection accuracy score (DetA) is obtained. In the process of tracking, if a method divides the trajectory of an instance into multiple segments or reflects a continuous trajectory in multiple instances, a lower correlation accuracy score (AssA) will be obtained. At the same time, additional popular indicators including DetRe, DetPr, AssRe, AssPr, LocA and IDS are also used for methods evaluation.
\subsection{Implementation Details}
The network is pre-trained on the COCO dataset \cite{lin2014microsoft} and then fine-tuned on the new dataset, using ResNet-101 \cite{he2016deep} as the backbone network structure. Based on Pytorch, the model is trained end-to-end in 12 epochs with 4 NVIDIA Titan X GPUs and implemented on MMDetection \cite{chen2019mmdetection}. In this process, the Adam optimizer \cite{kingma2014adam} is set to an initial learning rate of 0.0002, warm-up stage is used for the first 500 iterations, and the learning rate is attenuated by 0.1 after 9 epochs and 11 epochs. Through subsequent experimental verification, we select 11 feature layers in the backbone network to extract instance features, and the feature embedding length $e$ is 352. The maximum number of identifiers ${{N}_{m}}$, the maximum interval ${{T}_{m}}$ and the maximum track storage time $\tau$ of a single video sequence are set to 50, 10 and 30, respectively.
\subsection{Main Results}
\begin{figure*}
\centerline{\raisebox{0.08\columnwidth}{(a)}
  \subfloat{\includegraphics[width=0.18\textwidth]{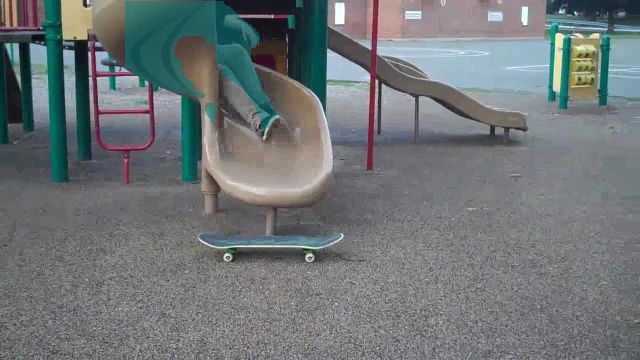}}\
  \subfloat{\includegraphics[width=0.18\textwidth]{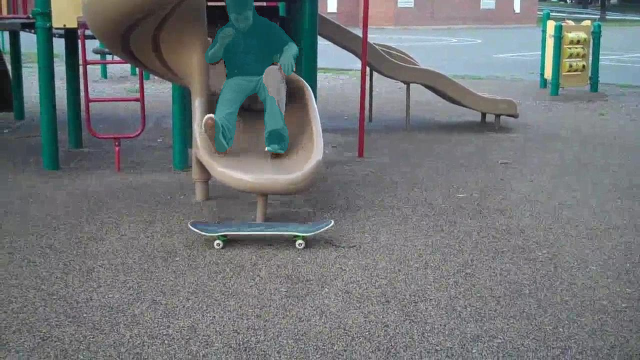}}\ 
  \subfloat{\includegraphics[width=0.18\textwidth]{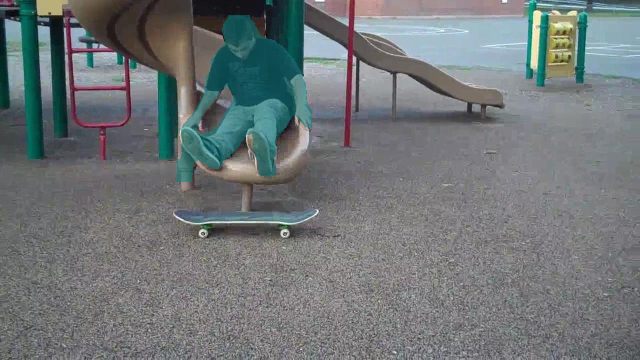}}\ 
  \subfloat{\includegraphics[width=0.18\textwidth]{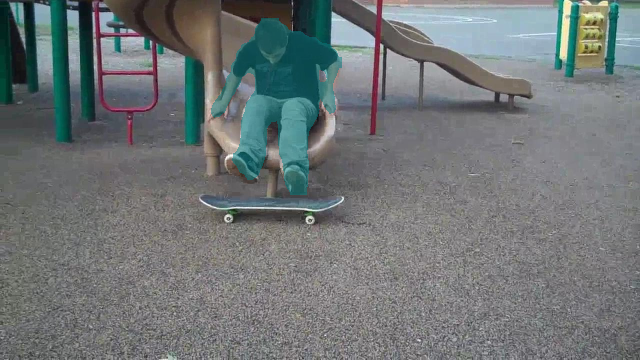}}\ 
  \subfloat{\includegraphics[width=0.18\textwidth]{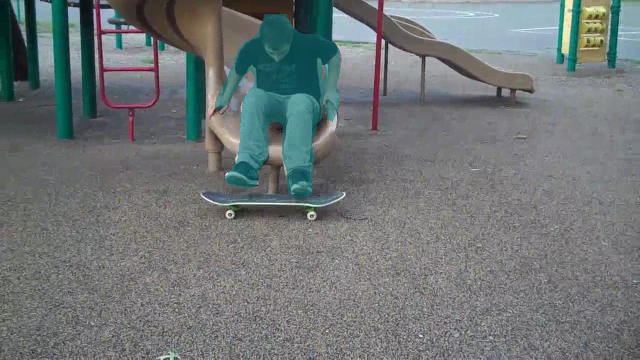}}}
\vspace{-0.1in}
\centerline{\raisebox{0.08\columnwidth}{(b)}
  \subfloat{\includegraphics[width=0.18\textwidth]{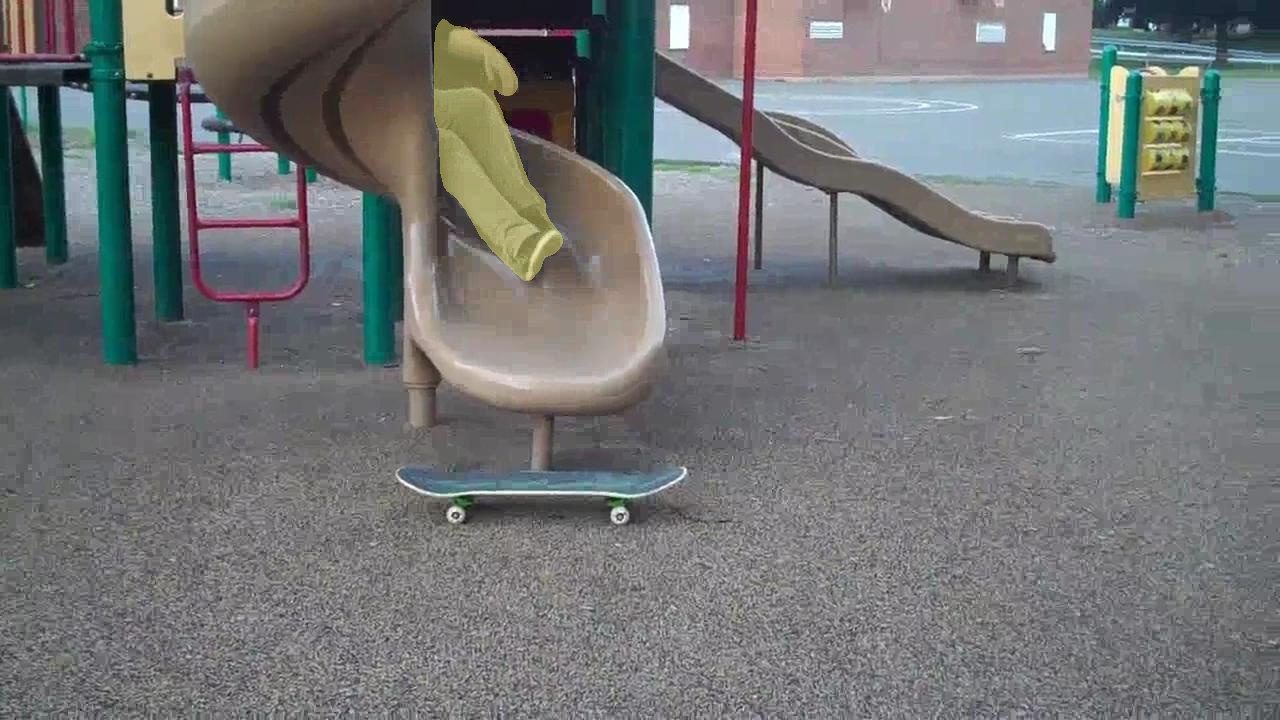}}\
  \subfloat{\includegraphics[width=0.18\textwidth]{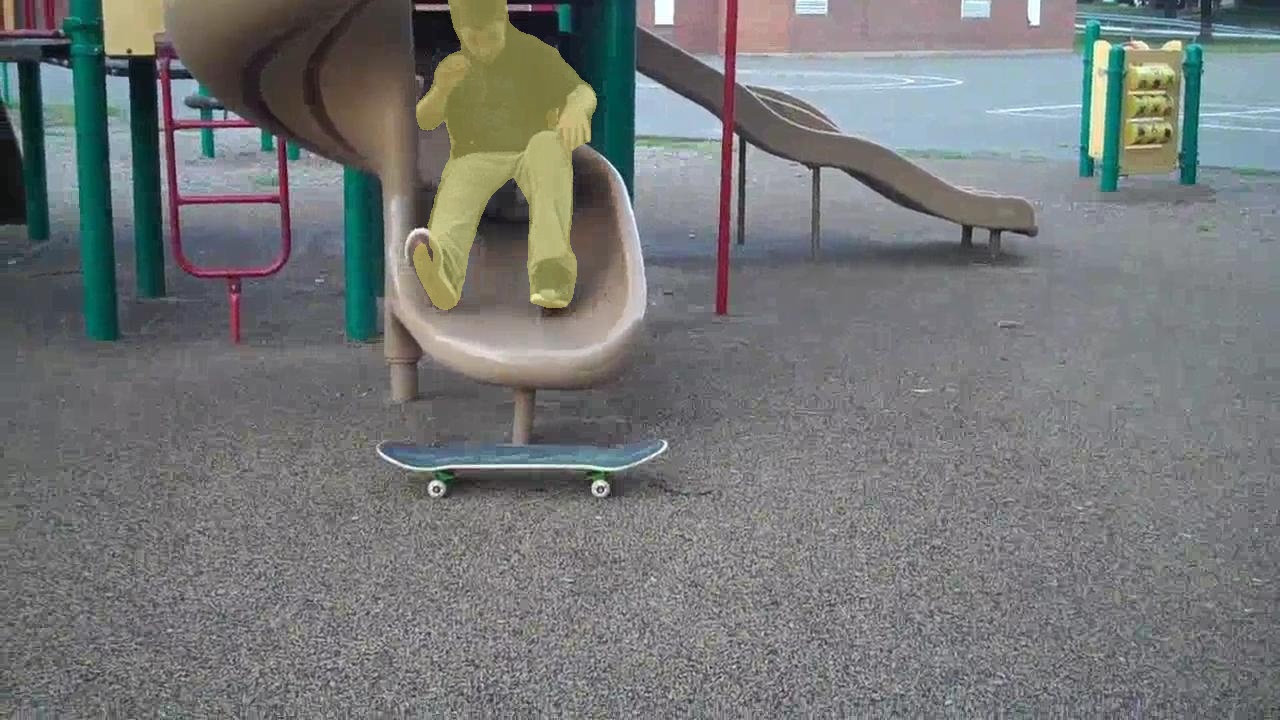}}\ 
  \subfloat{\includegraphics[width=0.18\textwidth]{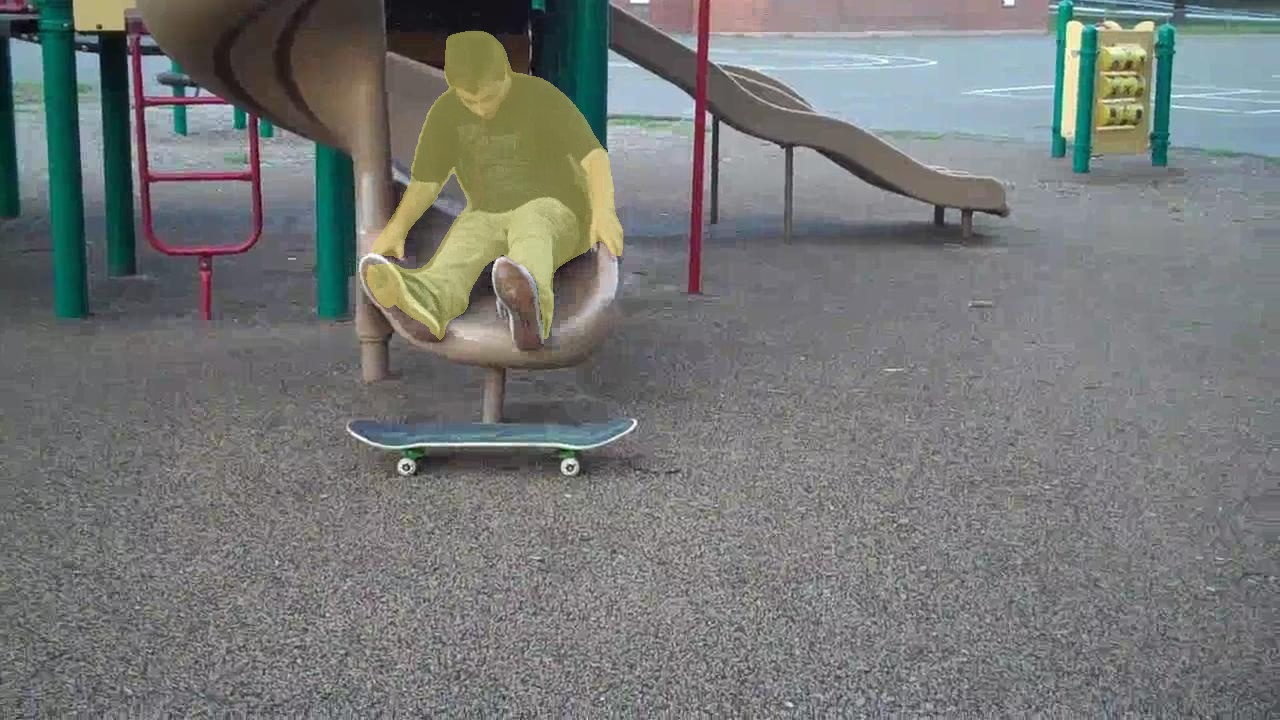}}\ 
  \subfloat{\includegraphics[width=0.18\textwidth]{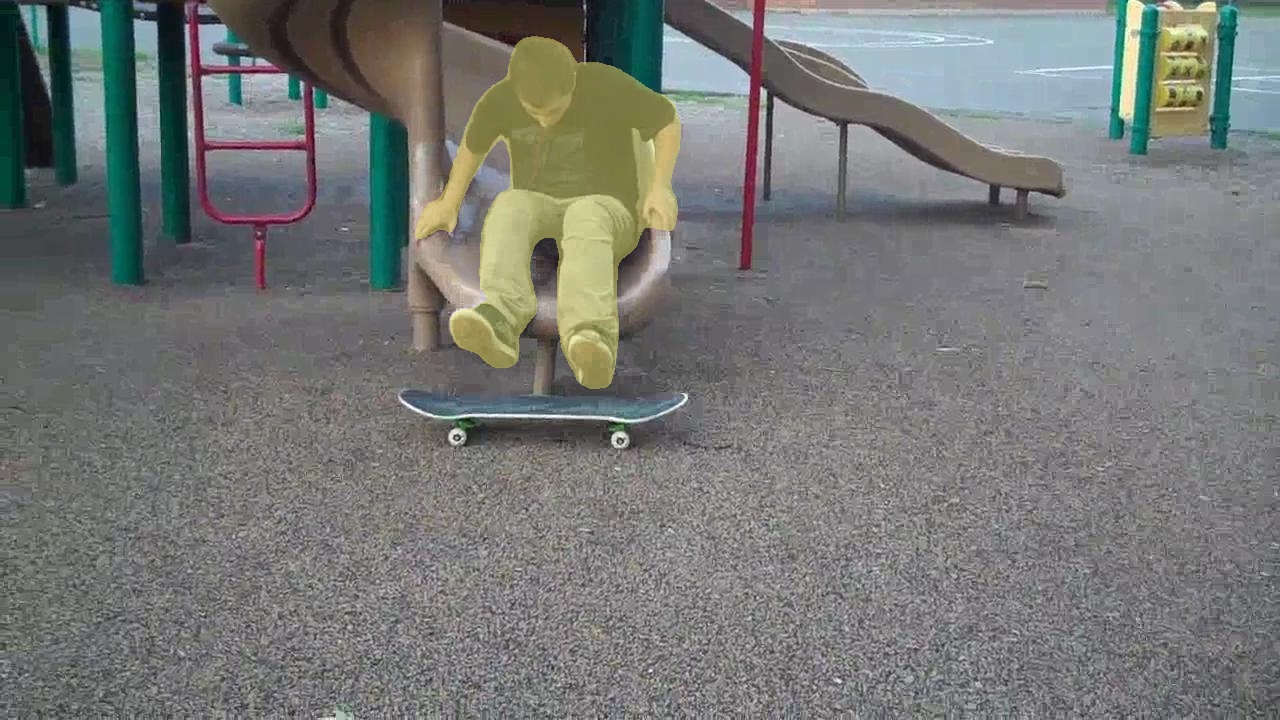}}\ 
  \subfloat{\includegraphics[width=0.18\textwidth]{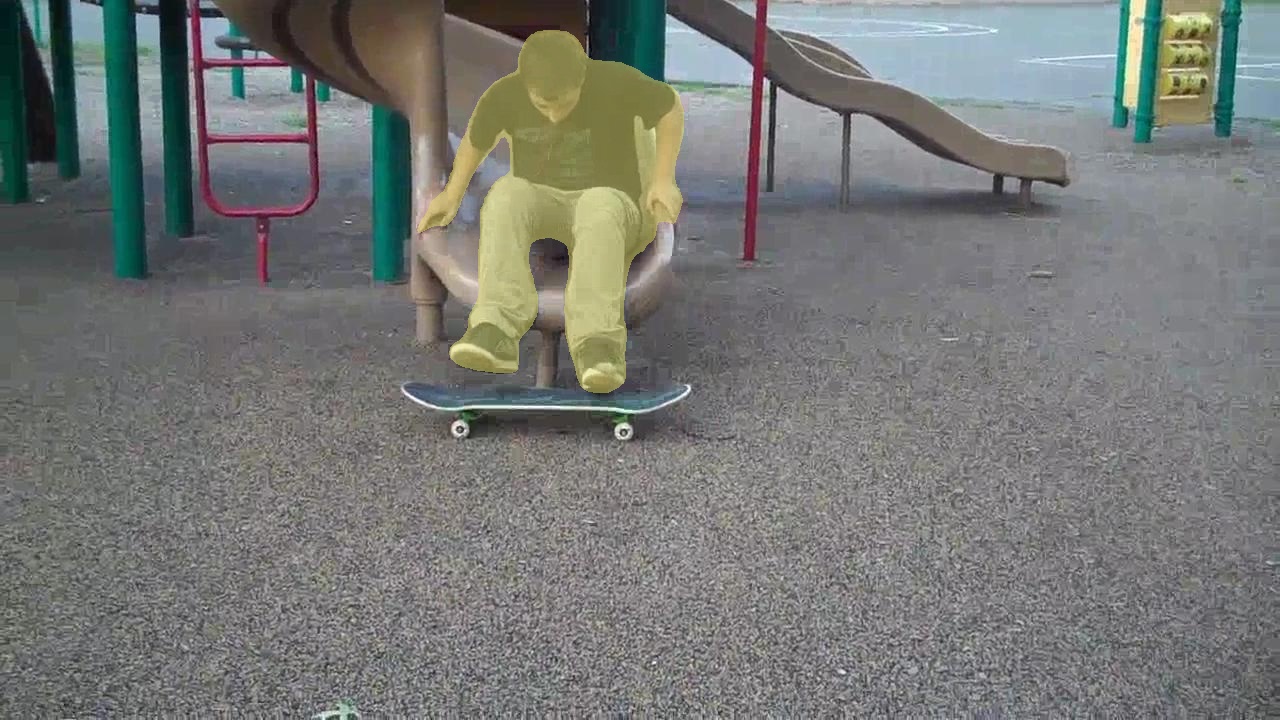}}}
\vspace{-0.1in}
\centerline{\raisebox{0.08\columnwidth}{(c)}
  \subfloat{\includegraphics[width=0.18\textwidth]{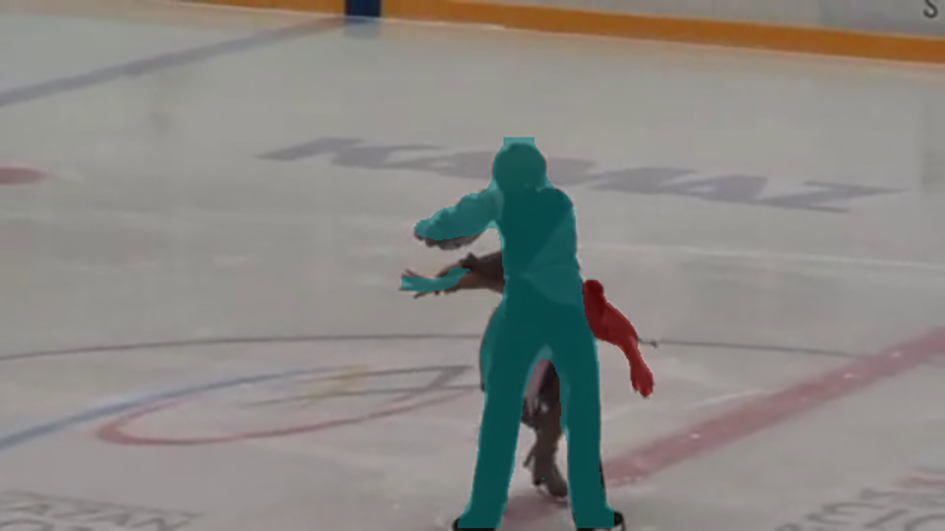}}\
  \subfloat{\includegraphics[width=0.18\textwidth]{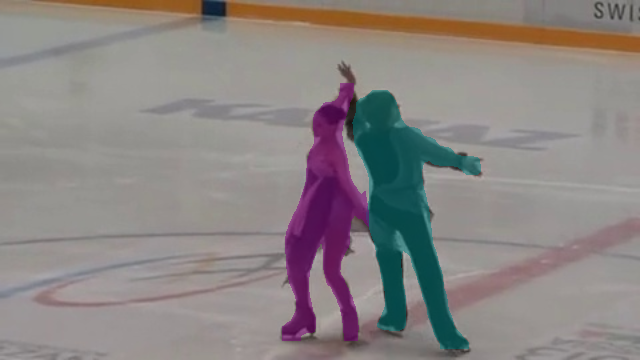}}\ 
  \subfloat{\includegraphics[width=0.18\textwidth]{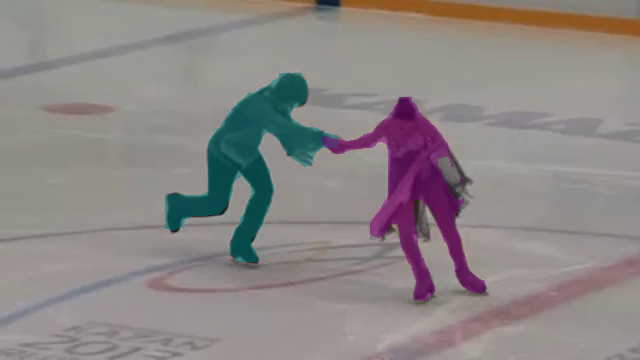}}\ 
  \subfloat{\includegraphics[width=0.18\textwidth]{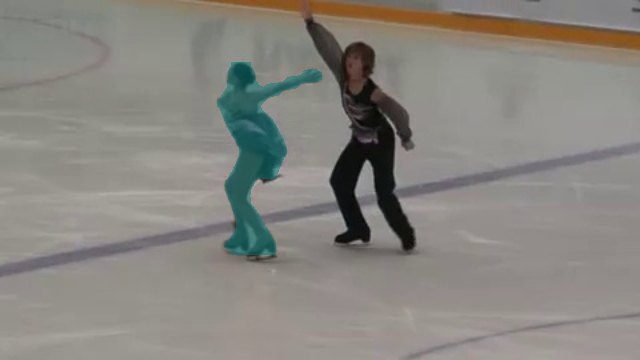}}\ 
  \subfloat{\includegraphics[width=0.18\textwidth]{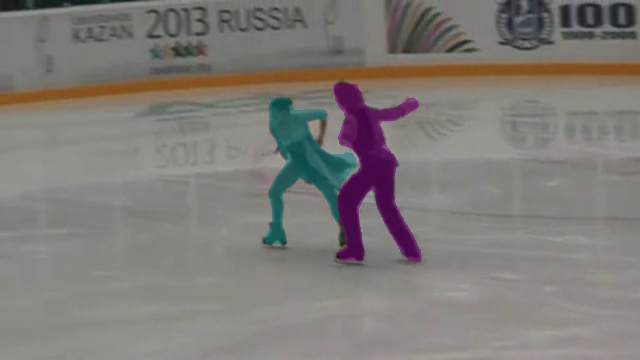}}}
\vspace{-0.1in}
\centerline{\raisebox{0.08\columnwidth}{(d)}
  \subfloat{\includegraphics[width=0.18\textwidth]{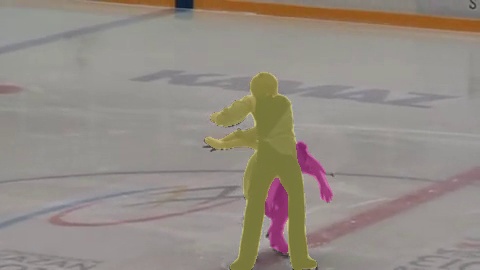}}\
  \subfloat{\includegraphics[width=0.18\textwidth]{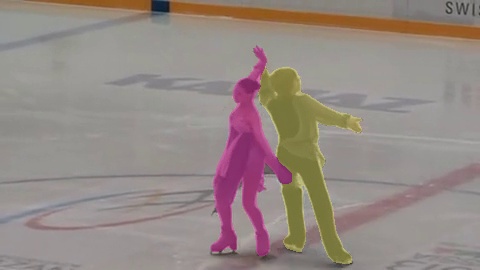}}\ 
  \subfloat{\includegraphics[width=0.18\textwidth]{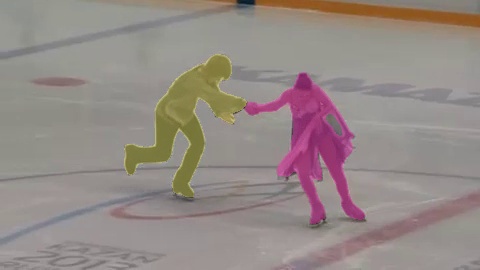}}\ 
  \subfloat{\includegraphics[width=0.18\textwidth]{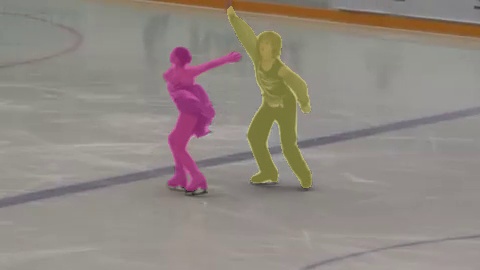}}\ 
  \subfloat{\includegraphics[width=0.18\textwidth]{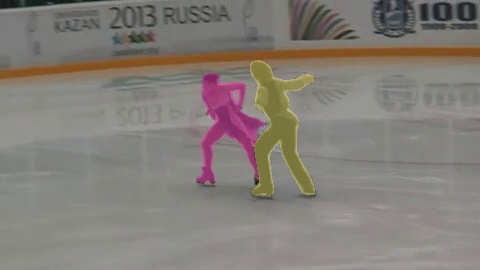}}}
\vspace{-0.1in}
\centerline{\raisebox{0.08\columnwidth}{(e)}
  \subfloat{\includegraphics[width=0.18\textwidth]{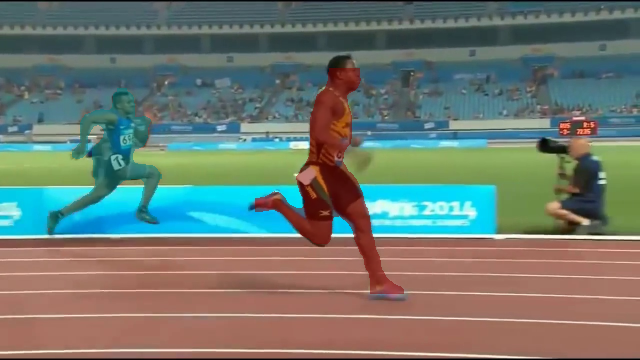}}\
  \subfloat{\includegraphics[width=0.18\textwidth]{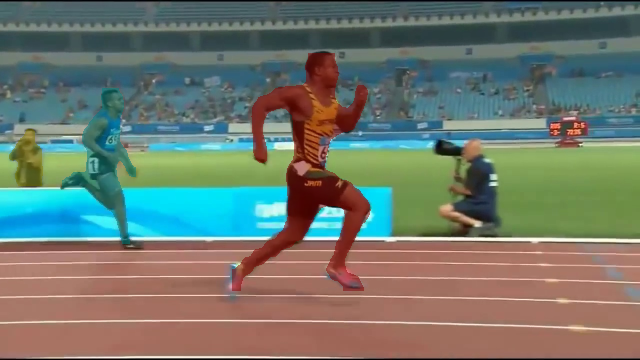}}\ 
  \subfloat{\includegraphics[width=0.18\textwidth]{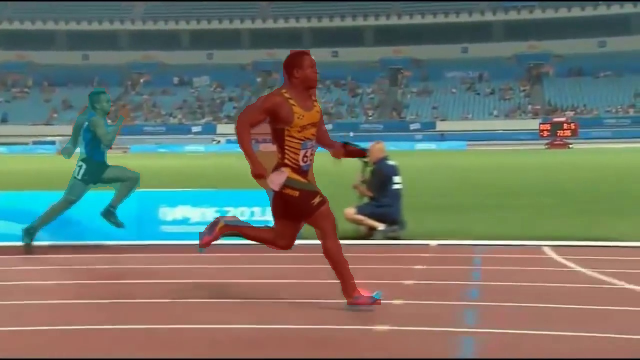}}\ 
  \subfloat{\includegraphics[width=0.18\textwidth]{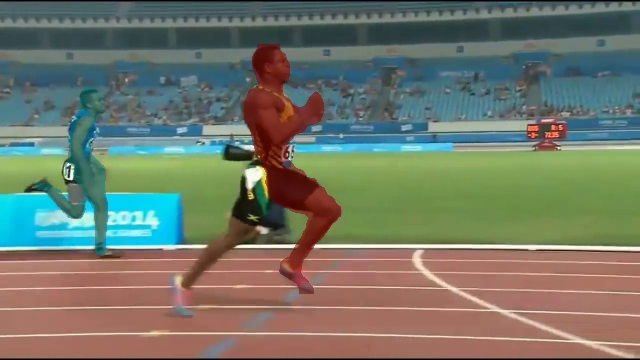}}\ 
  \subfloat{\includegraphics[width=0.18\textwidth]{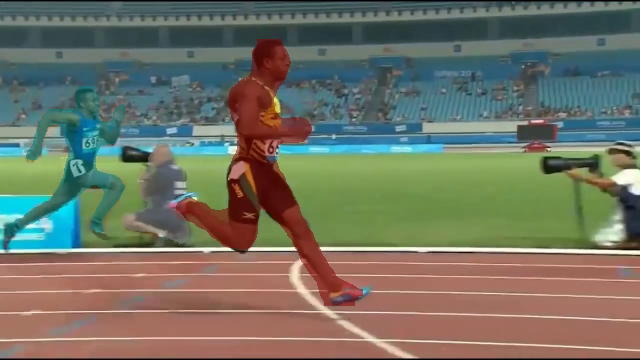}}}
\vspace{-0.1in}
\centerline{\raisebox{0.08\columnwidth}{(f)}
  \subfloat{\includegraphics[width=0.18\textwidth]{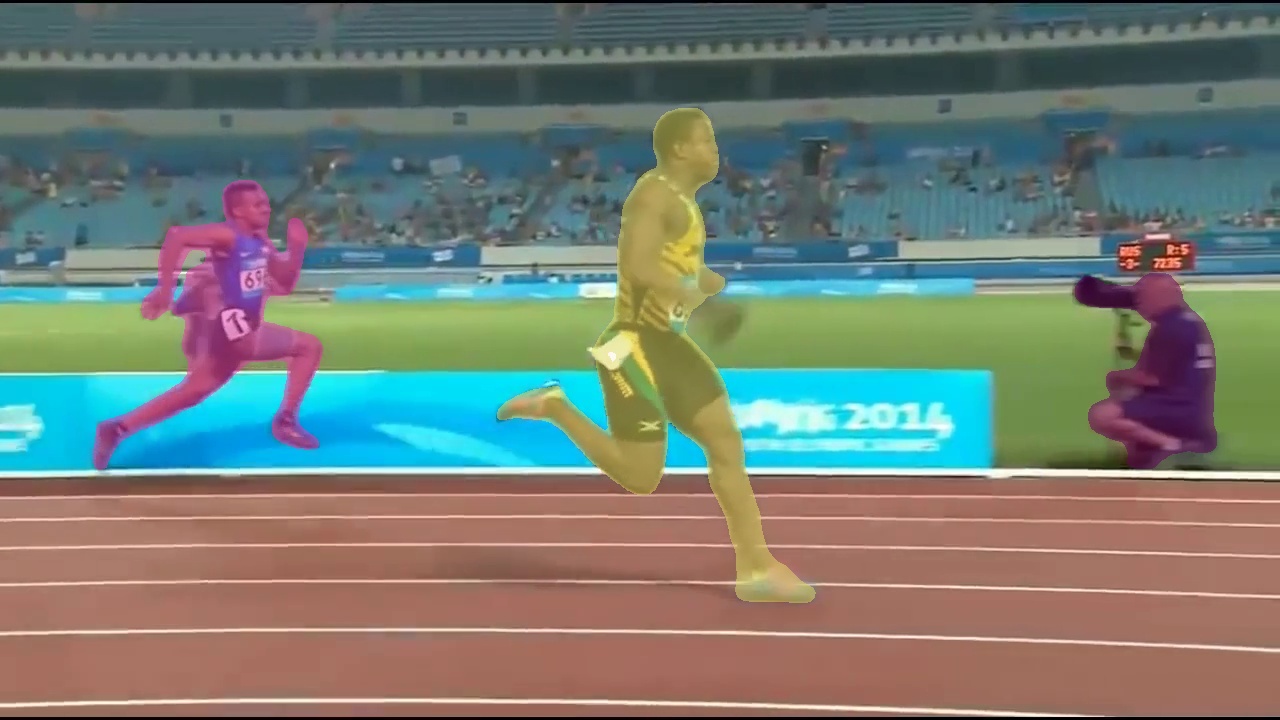}}\
  \subfloat{\includegraphics[width=0.18\textwidth]{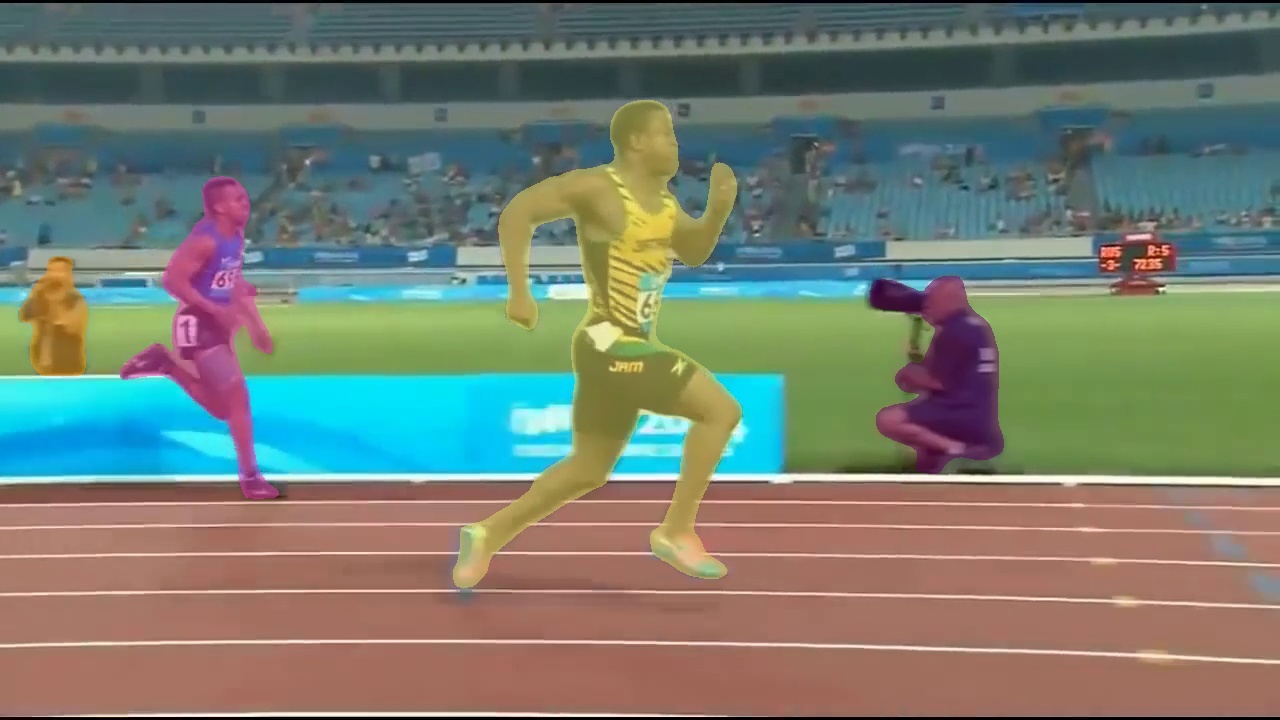}}\ 
  \subfloat{\includegraphics[width=0.18\textwidth]{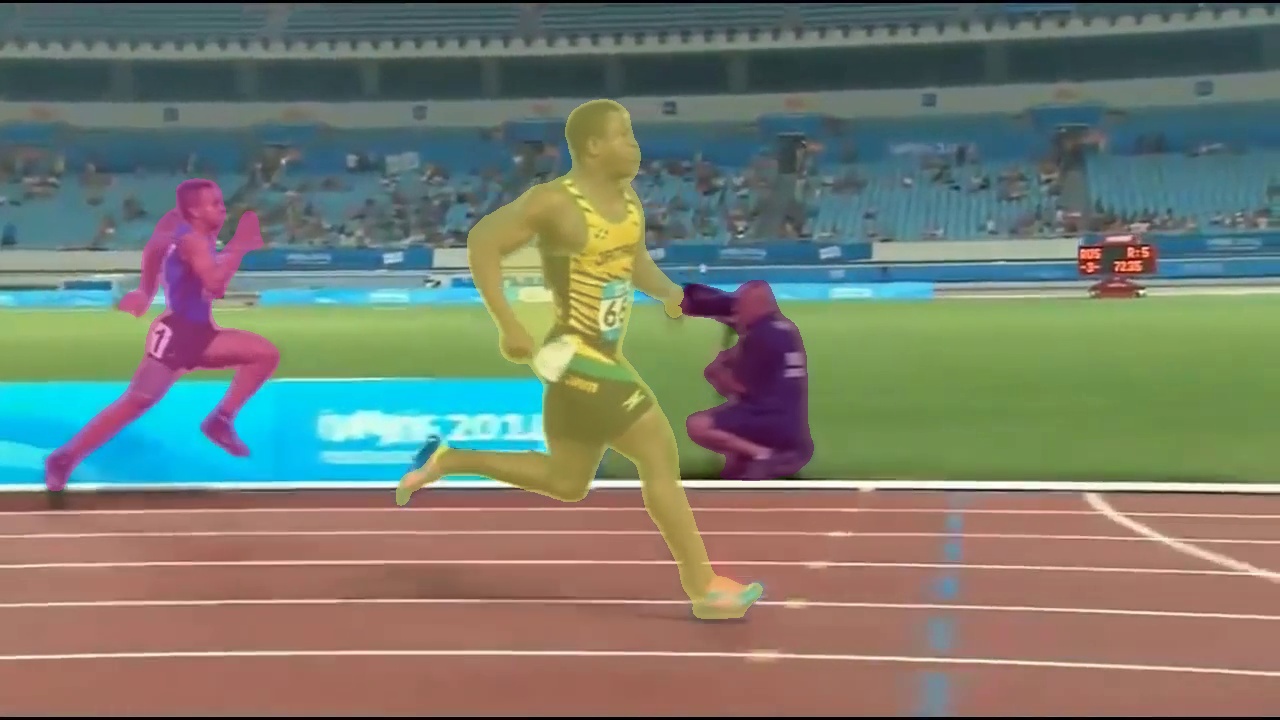}}\ 
  \subfloat{\includegraphics[width=0.18\textwidth]{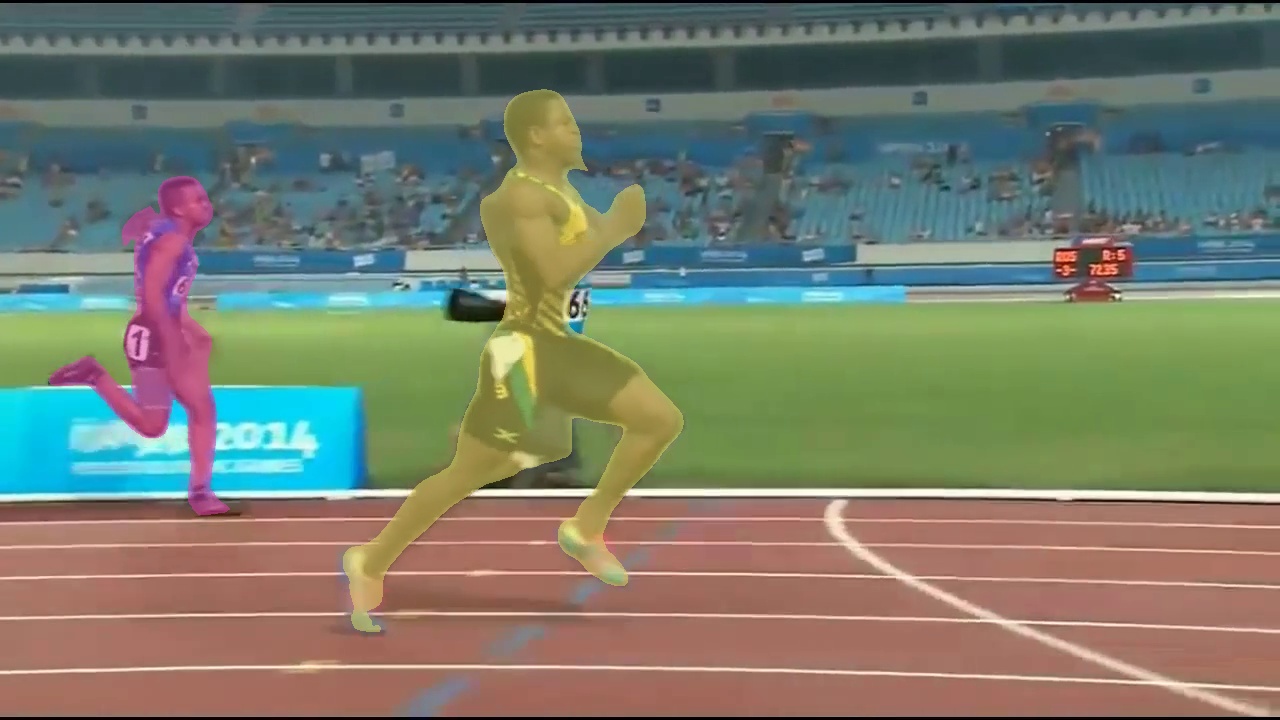}}\ 
  \subfloat{\includegraphics[width=0.18\textwidth]{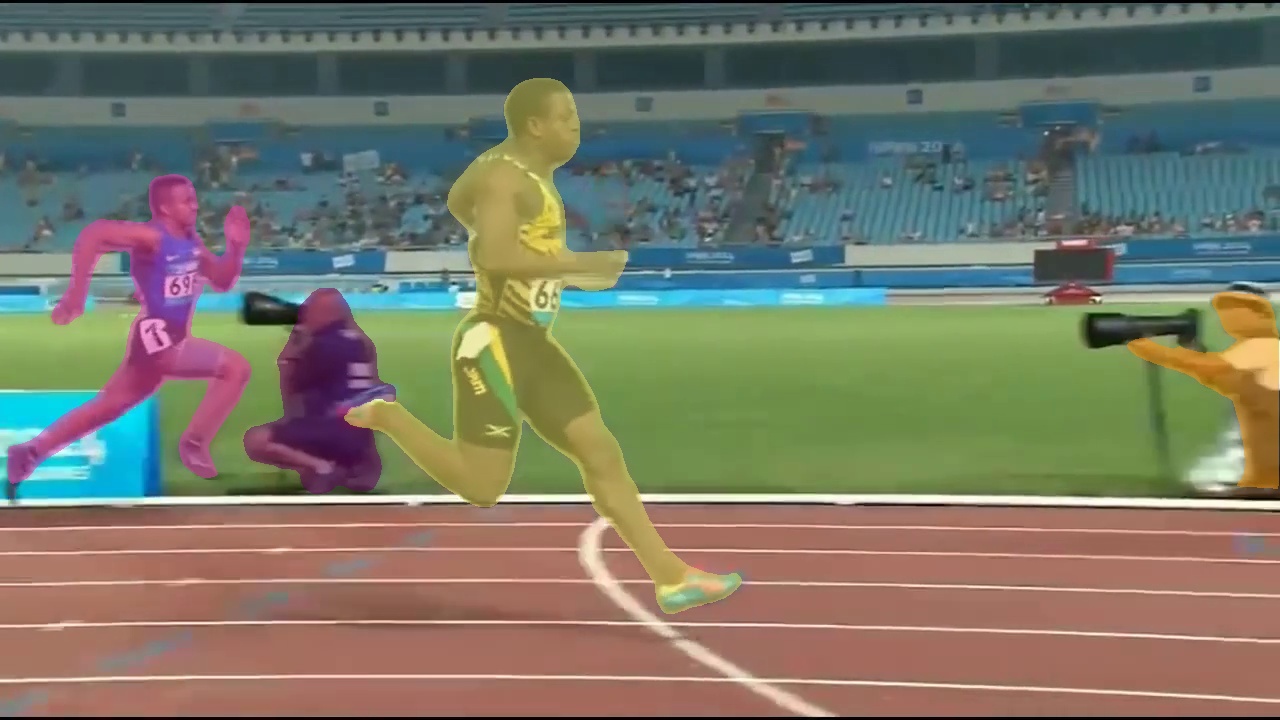}}}
\setcounter{subfigure}{0}
\caption{Comparison of segmentation results between MaskTrack-RCNN and the method proposed in this paper. Each line is the segmentation results of 5 frames from PVIS test set, and the same color is used to represent the same instance.}
\label{f2}
\end{figure*}

In the experimental process, the proposed method is compared with several state-of-the-art methods, such as MaskTrack-RCNN \cite{yang2019video} and SipMask, on PVIS dataset test set. The visualization results after comparison are shown in Fig. \ref{f2} with the specific quantitative results shown in TABLE \ref{t1}, which can illustrate the superiority of the proposed method in many aspects.

In Fig. \ref{f2}, Line (a), (c), (e) are the results of Masktrack-RCNN. and line (b), (d), (f) are the results of the proposed method on the same video sequences. Since the proposed method extracts instance feature embeddings on the multi-scale feature maps in the feature extraction module and includes three association stages in the matching process, it can better cope with the situation that the instance is partially blocked in Line (b) and the one that the instance's identity cannot be determined when the two instances are close to each other in Line (d). In contrast, MaskTrack-RCNN is affected in video sequences with occlusion and thus cannot segment instances accurately. The instance identity will be exchanged after two instances are interleaved, which cannot guarantee the consistency of the same instance identity in the same video sequence. 

Compared with two baselines, the proposed method performs best in terms of the overall video instance segmentation effects according to TABLE \ref{t1}, as there is little difference in the performance of these methods in the video with a single instance and slow motion condition. However, in the video with more characters, occlusion and high-speed motion, the proposed method with the instance features can capture more semantic information have better tracking performance, which alleviates the ID-switch problems caused by the rapid change of instance size in the process of high speed motion.
\begin{table}[!htp]
\centering
    \begin{tabular}{c|c|cccc}
\toprule[1pt]
    Methods& Type & HOTA &DetA &AssA  & IDS \\ 
\midrule[1pt]
    MaskTrack-RCNN \cite{yang2019video}&Two-stage&	54.3&	42.0&	72.8&	91\\
    SipMask \cite{cao2020sipmask}&One-stage&	51.9&	38.7&	72.2	&76\\
Ours	&One-stage&	58.7&	45.8&	77.2&	65 \\ 
\bottomrule[1pt]
    \end{tabular}
\caption{Quantitative evaluation results on a test set of the PVIS dataset compared with state-of-the-art video instance segmentation methods.}
\label{t1}
\end{table}
\subsection{Ablation Study}
\textbf{Feature embeddings:}

Since our method expands the instance tracking module based on SOLO \cite{wang2020solo}, the feature embeddings obtained by concatenating the features of the same instance at each layer in FPN \cite{lin2017feature} will conflict with each other. In other words, features of different sizes of the same instance represent different instances in SOLO. Direct concatenation will lead to serious id-switch problems. To handle this problem, we focus on the last four stages of ResNet-101 \cite{he2016deep} for feature embeddings and extract 2, 2, 5 and 1 feature layers evenly according to the proportion of the number of blocks in each stage, that is, 11 feature layers were extracted from layers 7, 10, 16, 22, 34, 46, 58, 70, 82, 91 and 100 of the backbone networks. At the same time, we halve the number of feature layers and select 6 feature layers 7, 16, 34, 58, 82, 100 for a total of 6 feature layers as a comparison. In TABLE \ref{t2}, the representation effects of instance features obtained by selecting different number of feature layers are compared. Meanwhile, in order to observe the influence of embedding length on appearance association effect, the number of feature layers is fixed, and the output length of each feature layer is doubled. The corresponding results of embedding features of different lengths are shown in TABLE \ref{t2}.
\begin{table}[!htp]
\newcommand{\tabincell}[2]{\begin{tabular}{@{}#1@{}}#2\end{tabular}}
\centering
    \begin{tabular}{c|c|cccc}
\toprule[1pt]
\tabincell{c}{Embedding\\length}	&\tabincell{c}{Featuremap\\layers}&	HOTA&	DetA&	AssA&	FPS \\ 
\midrule[1pt]
16&	11&	57.7&	45.5&	75.3&	4.0\\
32&	11&	\textbf{58.7}&	45.8&	\textbf{77.2}&	4.2\\
64&	11&	57.7&	45.8&	75.0&	3.0\\
32&	6&	57.9&	\textbf{46.2}&	75.0&	4.2\\
64&	6&	58.2&	46.1&	75.4&	\textbf{4.4}\\
\bottomrule[1pt]
    \end{tabular}
\caption{The length of a single-layer feature embedding and the number of layers which extracts features are evaluated on the test set of the PVIS dataset. The best results for each column are shown in \textbf{bold}.}
\label{t2}
\end{table}

By comparing the performance based on different number of feature layers as in TABLE \ref{t2}, we can observe that when the number of feature layers is 6, increasing the length of feature output of each layer cannot significantly enhance the performance effect of the instance. AssA only increases from 75.0\% to 75.4\%, and DetA slightly decreases from 46.2\% to 46.1\%, which is different from the case when the number of feature layers is 11. The results suggest that increasing the number of feature layers can capture more instance information in different scales, thereby improving the recognition ability of acquired features and facilitating the subsequent instance identity matching using appearance features. In addition, it is observed that the detection accuracy increases slightly when the number of feature layers is small, which are mainly attributed to the conflict between the instance segmentation of SOLO \cite{wang2020solo} and the matching module added. The balance between the segmentation accuracy and tracking accuracy is maintained by setting the number of feature layers to 11 in the experiment.

In addition, it can also be seen when the number of feature layers is fixed at 11 and the length does not reach the high limitation, the representation effect of the instance is enhanced with the increase of embedded feature length. Specifically, the association effect is significantly increased by 1.9\%. In terms of segmentation, due to the lack of adjustment of segmentation results according to the tracking results, only the instance segmentation is enhanced in part of the videos with occlusion by 0.3\% improvement. With the increase of the embedding length, the requirements for memory and computation power are also increased, but the inference speed is further reduced. Considering the trade-off between efficiency and accuracy, the single-layer embedding length and the number of feature layer is set to 32 and 11, respectively.

\textbf{Sampling Strategy:}

\begin{figure}
\centerline{
\subfloat{\includegraphics[width=0.23\textwidth]{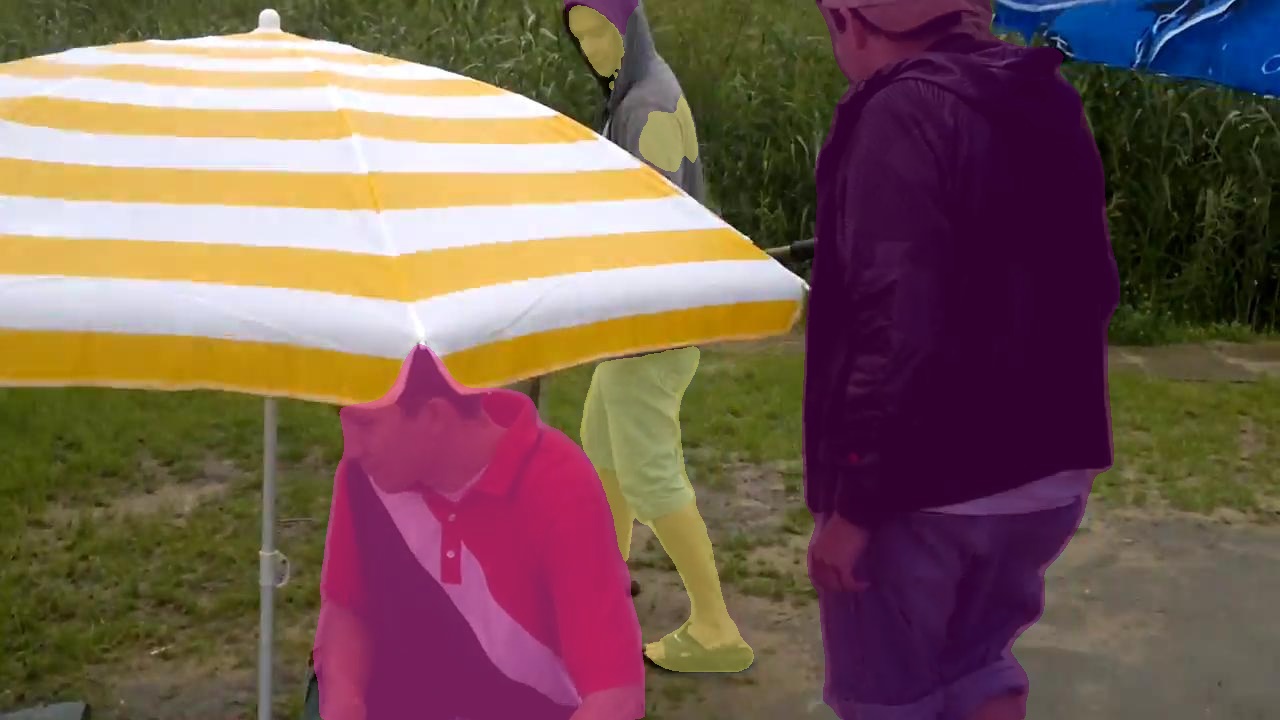}}\
\subfloat{\includegraphics[width=0.23\textwidth]{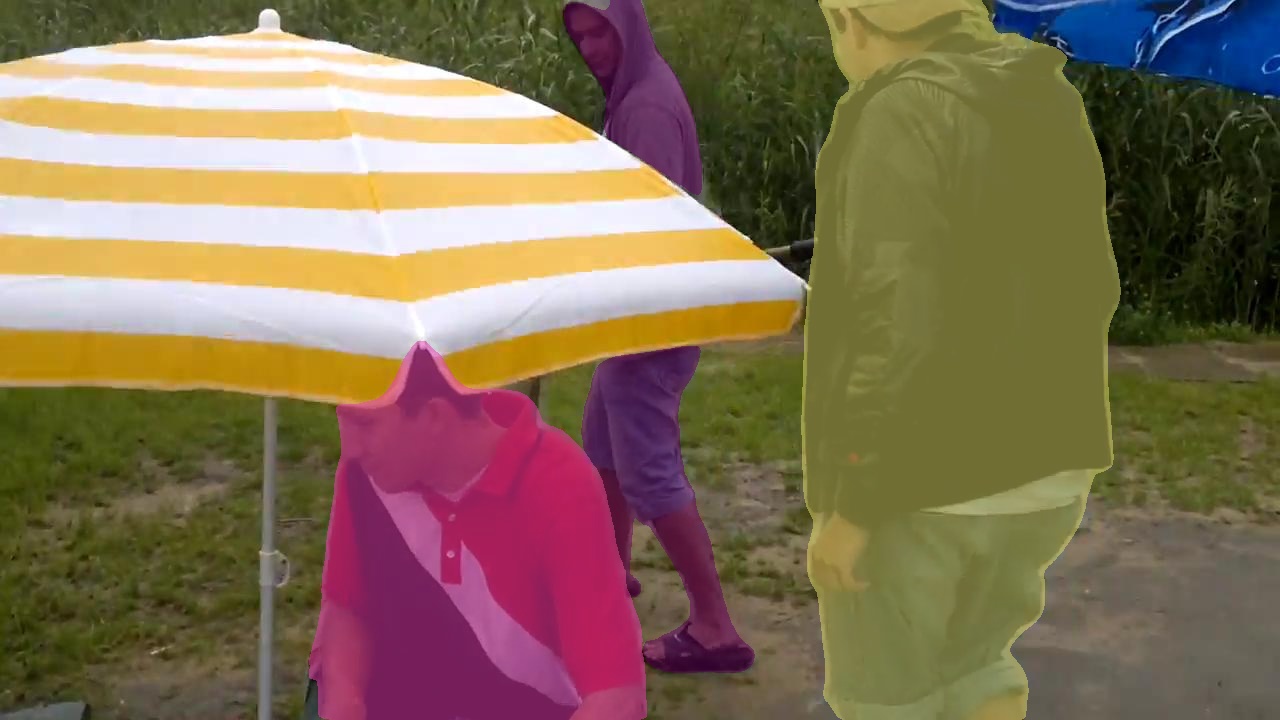}}}
\vspace{-0.1in}
\centerline{
\subfloat{\includegraphics[width=0.23\textwidth]{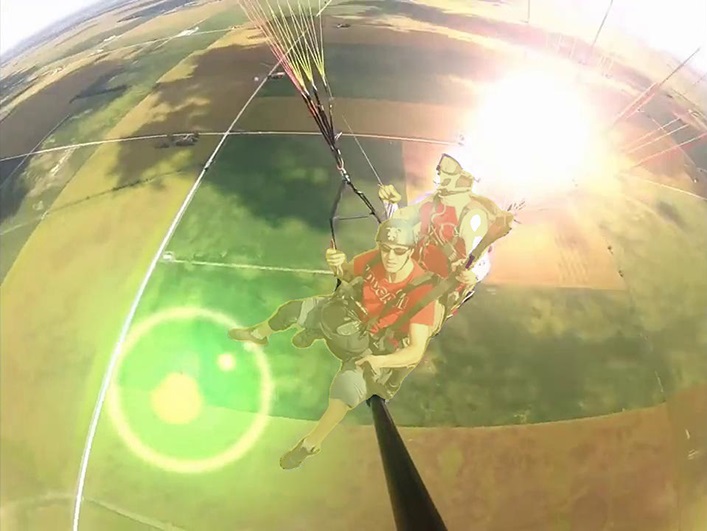}}\
\subfloat{\includegraphics[width=0.23\textwidth]{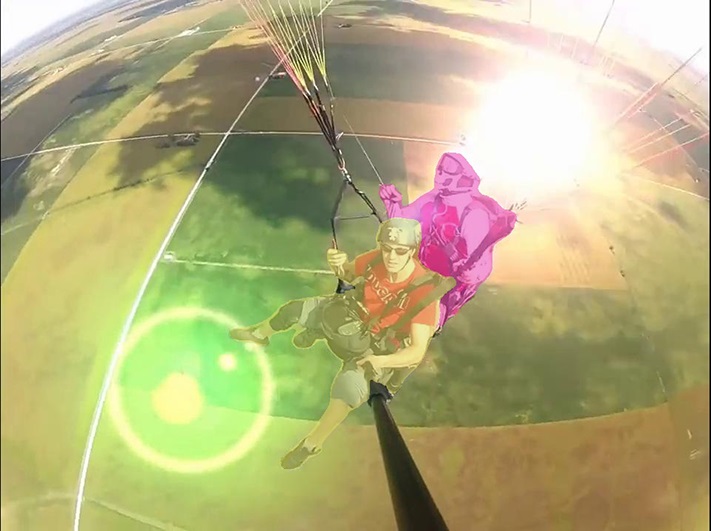}}}
\setcounter{subfigure}{0}
\caption{Comparison of segmentation results between two sampling strategies. The bounding box centroid sampling strategy is used on the left side, and the centroid sampling strategy for maximum area contour is used on the right side.}
\label{f3}
\end{figure}

To verify the effectiveness of the proposed centroid sampling strategy for the maximum area contour, the strategy is compared with the bounding box centroid sampling strategy. The bounding box centroid sampling strategy of the target frame will lead to severe oscillation of the center point in some extreme action changes of instances and great fluctuation of the instance features in corresponding positions, which will affect the subsequent appearance association. In addition, in the case of large overlap area such as the lower portion of Fig. \ref{f3}, the bounding box centers of the two figures are close to each other or overlap. In this case, the bounding box centroid sampling strategy will lead to similar embedded features extracted by the two instances, and the indistinguishable degree will lead to the fuzzy positive sample problem. The proposed sampling strategy can discard the overlapping part and the smaller area after the instance is segmented and the maximum area mask centroid is selected to extract features. Fig. \ref{f3} shows that under the interference of occlusion and overlapping figures, the proposed sampling method can correctly extract the features of corresponding instances and alleviate the fuzzy positive sample problem.  In TABLE \ref{t3}, HOTA increases 0.9\% and less id-switch problems occur.
\begin{table}[!htp]
\newcommand{\tabincell}[2]{\begin{tabular}{@{}#1@{}}#2\end{tabular}}
\centering
    \begin{tabular}{c|cccc}
\toprule[1pt]
Sampling strategy&	HOTA&	DetA&	AssA&	FPS \\ 
\midrule[1pt]
Center of bbox&	57.8&	45.2&	76.2	&88\\
Centroid for maximum area contour	&58.7&	45.8&	77.2&	65\\
\bottomrule[1pt]
\end{tabular}
\caption{The two feature sampling strategies are compared and evaluated on the test set of PVIS dataset. }
\label{t3}
\end{table}

\textbf{Data Association Method:}

In this paper, data association is mainly divided into three stages: Kalman filter \cite{maybeck1990kalman}, IOU association and feature embedding association. Mask IOU is used to replace bounding box IOU in the experiment with TABLE \ref{t4} showing the comparison results of each matching method. It can be observed from the results that appearance matching by feature is the main association method, and there is little difference between Kalman filtering or IOU association alone on this basis. However, using Kalman filter alone can remove a part of unreasonable trajectory and reduce the number of ID switches. Using mask IOU alone can obtain higher AssA than using bounding box IOU alone. The possible behind reason is that when some different action changes occur in the detection task, the coincidence degree of appearance contour can better reflect changes, thus reducing the number of ID switches. When the three correlation methods are used at the same time, they can deal with ordinary instance tracking, as well as the situation of fast movement or occlusion.
\begin{table}[!htp]
\newcommand{\tabincell}[2]{\begin{tabular}{@{}#1@{}}#2\end{tabular}}
\centering
    \begin{tabular}{c|c|cccc}
\toprule[1pt]
Motion module&	IOU&	HOTA&	DetA&	AssA&	IDS\\
\midrule[1pt]
Kalman	&None&	58.1&	45.8&	75.7&	91\\
None&	Bbox IOU	&58.1&	45.7&	75.8&	114\\
None&	Mask IOU	&58.3&	45.8&	76.3&	107\\
Kalman&	Mask IOU&	58.7&	45.8&	77.2	&65\\
\bottomrule[1pt]
    \end{tabular}
\caption{Different data association methods were compared and evaluated on test set of PVIS datasets.}
\label{t4}
\end{table}
\section{Conclusion}
In this paper, we develop a VIS method by involving feature extraction and data association modules on a single-stage detector. In order to improve the ID switches problem caused by huge size changes in the video, we select different scale feature layers of the backbone network to extract instance features and get multi-scale representation of the instance; to handle the occlusion problem or the overlap of sampling points due to close instances in the video, the centroid sampling strategy for maximum area contour is used to eliminate the effect. All these can promote the robustness of the proposed model for detection and tracking. In addition, due to the current lack of human-specific datasets, we obtained a new human dataset based on several public datasets to evaluate the effect of video human instance segmentation. The results on new dataset show that our method can achieves state-of-the-art performance as well as the robustness for complex video scenes.
\section*{Acknowlegement}
This work is partially supported by National Key Research and Development Program of China (2019YFC1521300), supported by National Natural Science Foundation of China (61971121, 61672365) and also supported by the Fundamental Research Funds for the Central Universities of China (JZ2019HGPA0102).
\bibliographystyle{IEEEtran}
\bibliography{vis}

\end{document}